\documentclass[sigconf]{acmart}
\AtBeginDocument{%
  }

\setcopyright{acmlicensed}
\copyrightyear{2026}
\acmYear{2026}
\acmDOI{XXXXXXX.XXXXXXX}
\acmConference[SIGIR '26]{the 49th International ACM SIGIR Conference on Research and Development in Information Retrieval}{July 20--24,
  2026}{Naarm, Australia}

\acmISBN{978-1-4503-XXXX-X/2018/06}

\usepackage{multirow}
\usepackage{tabularx}
\usepackage[ruled,vlined,linesnumbered]{algorithm2e}
\usepackage{amsmath}
\begin{document}

\title {TimelineReasoner: Advancing Timeline Summarization with Large Reasoning Models}

\author{Liancheng Zhang}

\affiliation{%
  \institution{Gaoling School of Artificial Intelligence, Renmin University of China}
  \city{Beijing}
  \country{China}
}
\email{zhangliancheng227@ruc.edu.cn}

\author{Xiaoxi Li}
\affiliation{%
  \institution{Gaoling School of Artificial Intelligence, Renmin University of China}
  \city{Beijing}
  \country{China}}
\email{xiaoxi_li@ruc.edu.cn}

\author{Zhicheng Dou}
\affiliation{%
  \institution{Gaoling School of Artificial Intelligence, Renmin University of China}
  \city{Beijing}
  \country{China}
}
\email{dou@ruc.edu.cn}

\renewcommand{\shortauthors}{Zhang et al.}
\settopmatter{printacmref=false} 

\begin{abstract}
The proliferation of online news poses a challenge to extracting structured timelines from unstructured content. 
While recent studies have shown that Large Language Models (LLMs) can assist {Timeline Summarization (TLS)}, these approaches primarily treat models as passive generators. The emergence of Large Reasoning Models (LRMs) presents an opportunity to reason over events actively, enabling iterative evidence acquisition, the detection of missing events, and the validation of temporal consistency. 
To systematically leverage the reasoning capabilities of LRMs, we propose \textbf{TimelineReasoner}, a novel framework that shifts TLS from static generation to an active, reasoning-driven process. 
Unlike prior work, TimelineReasoner adopts a two-stage framework: {Global Cognition}, which tracks events at a macroscopic level and continuously updates a global event memory, and {Detail Exploration}, which identifies informational gaps and refines the timeline via targeted document retrieval. To support this, TimelineReasoner incorporates several specialized mechanisms, including an Event Scraper for retrieving temporal event descriptions, a Timeline Updater for refining the timeline, and a Supervisor for detecting gaps in the timeline and guiding retrieval. Experimental results on open-domain TLS datasets demonstrate that TimelineReasoner significantly outperforms existing LLM-based TLS methods in terms of timeline accuracy, coverage, and coherence. On closed-domain TLS datasets, our method performs on par with or exceeds state-of-the-art approaches. This work not only pushes the boundaries of TLS but also highlights the broader potential of LRM-based reasoning frameworks for timeline summarization.

\end{abstract}




    

\maketitle

\section{Introduction}
\begin{figure*}
  \centering
  \includegraphics[width=0.8\textwidth]{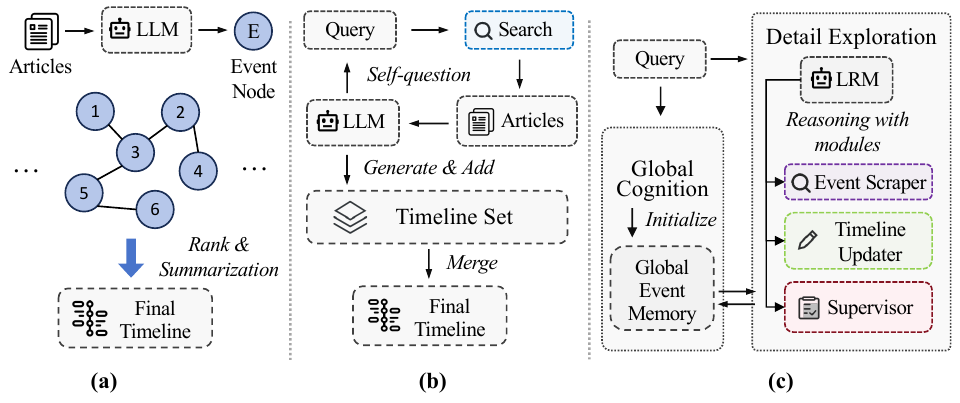}

  \caption{Comparison of methods using LLMs for TLS. (a) A method that uses LLMs to generate event nodes and constructs an event graph for timeline derivation via cluster ranking. (b) A framework that employs iterative self-questioning to link events, merging generated sub-timelines into a final output. (c) TimelineReasoner (Ours)\textbf{:} An active, reasoning-driven framework using LRMs. It maintains a dynamic global event memory during Global Cognition and employs an agentic loop for Detail Exploration to iteratively refine the timeline.}
  \label{fig:figure1}

\end{figure*}

In today's digital era, the rapid proliferation of news information presents a significant challenge: extracting coherent event narratives from an overwhelming volume of textual sources. Every minute, countless news articles are published, making it difficult for readers to track the evolution of complex events. \textit{Timeline Summarization (TLS)}~\cite{Allan2001TemporalSO, Chen2019LearningTA, Yu2021MultiTimeLineS, gholipour-ghalandari-ifrim-2020-examining} addresses this problem by distilling chronological event descriptions from large corpora, producing structured summaries that capture key developments over time. 

Recent advances in {Large Language Models (LLMs)}~\cite{Kojima2022LargeLM,Achiam2023GPT4TR,Bai2023QwenTR} have introduced new opportunities for TLS, leveraging their superior comprehension and generation capabilities to produce high-quality summaries. However, existing LLM-based approaches~\cite{hu-etal-2024-moments,Wu2025UnfoldingTH} primarily treat these models as passive generators, which means feeding them raw articles and directly generating timeline outputs. As illustrated in Figure~\ref{fig:figure1} (a), such methods typically construct an event graph and then generate the final timeline from it. 
While effective at extracting local details, these methods lack a global mechanism for dynamically acquiring and integrating event information across the entire timeline, making it difficult to ensure coherent coverage and temporal consistency. 
Retrieval and summarization are performed in isolation, without validating the logical continuity of events. As a result, timelines often suffer from fragmentation, redundancy, and temporal inconsistency.

Meanwhile, {Large Reasoning Models (LRMs)}~\cite{Contributors2024OpenAIOS,qwq32b,DeepSeekAI2025DeepSeekR1IR} have demonstrated remarkable problem-solving abilities in domains like mathematics, coding, and scientific research. Their capability for structured reasoning, and tool-augmented workflows makes them well-suited for tasks requiring multi-step analysis, which are precisely the demands of TLS. In particular, applying LRMs to TLS is promising, as their deep, multi-step reasoning capabilities can support dynamic evidence acquisition, missing-event discovery, and temporal consistency verification across long timelines. However, the field still lacks a principled framework that models timeline construction as a hierarchical cognitive process. Such a framework should move beyond passively generating timelines, and instead dynamically construct comprehensive and temporally accurate timelines through iterative reasoning and targeted information retrieval.

To address this, we propose \textbf{TimelineReasoner}, a reasoning-driven framework that actively acquires, integrates, and refines event information for TLS. TimelineReasoner leverages the capabilities of LRMs to dynamically construct timelines through iterative reasoning and targeted information retrieval. Our framework adopts a two-stage design—\textbf{Global Cognition} and \textbf{Detail Exploration}—supported by specialized mechanisms. 

In the Global Cognition stage, TimelineReasoner aims to establish an initial, coarse-grained understanding of the overall event landscape, identifying the major events and their approximate temporal structure to form a global view of the evolving news narrative. This stage is initialized through a broad, one-pass retrieval process that gathers representative news articles, from which high-level event information is extracted by the \textbf{Event Scraper}. The resulting global understanding serves as a foundation for subsequent fine-grained refinement, and is later updated using newly discovered evidence from the Detail Exploration stage.

In the Detail Exploration stage, TimelineReasoner focuses on identifying missing or incomplete events and acquiring additional relevant information to fill these gaps, ensuring both detail completeness and temporal consistency. It performs a search-interleaved reasoning process, analyzing the global event memory and the evolving timeline memory to identify informational gaps and generate targeted sub-queries to the Event Scraper for fine-grained event retrieval, and produce sub-timelines that are integrated into the timeline memory via the \textbf{Timeline Updater}. A \textbf{Supervisor} module further monitors coverage gaps, temporal inconsistencies, and informational deficiencies, guiding the LRM with structured search plans to iteratively refine the timeline until it achieves high completeness, coherence, and factual accuracy.

This reasoning-centric architecture enables TimelineReasoner to go beyond static summarization, actively bridging missing temporal links, resolving inconsistencies, and delivering structured, high-fidelity timelines from unstructured multi-document news corpora. 

In summary, our main contributions are as follows:

(1)  We introduce a coarse-to-fine reasoning paradigm for timeline construction that decouples \textit{global event comprehension} from \textit{local detail verification}. This paradigm formulates timeline summarization as an iterative reasoning process that progressively refines the timeline through targeted evidence acquisition, addressing the fundamental challenges of temporal inconsistency, incomplete coverage, and fragmented narratives in existing TLS methods.

(2) Building on this paradigm, we propose \textbf{TimelineReasoner}, a reasoning-driven framework that instantiates the coarse-to-fine process via modular components for event acquisition, timeline refinement, and consistency maintenance. The framework enables active interaction between reasoning and retrieval, allowing the timeline to be incrementally improved rather than passively generated in a single step.

(3) We design a \textit{dynamic timeline memory updating mechanism} that continuously integrates newly retrieved evidence into an evolving timeline while preserving temporal and factual consistency. This mechanism allows TimelineReasoner to revise and enrich previous timeline memory as new information becomes available.

(4) We introduce an \textbf{agentic supervision mechanism} for TLS, which performs meta-level analysis over the evolving timeline to identify coverage gaps, temporal inconsistencies, and informational deficiencies. This supervision mechanism generates structured search plans to guide TimelineReasoner toward more targeted retrieval and reasoning, thereby improving the completeness and coherence of the resulting timelines.

\begin{figure*}
  \centering
  \includegraphics[width=1\textwidth]{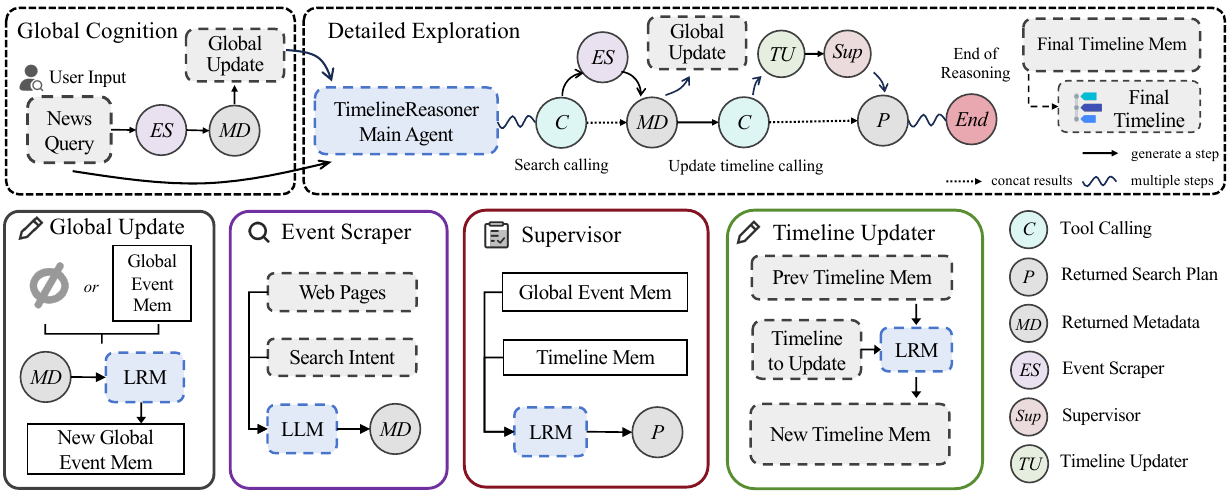}
  \caption{Framework of TimelineReasoner. It contains two stages: Global Cognition searches for the news query, uses an LRM to generate global event memory, and dynamically updates it in the later process. Detailed Exploration takes a main reasoning model as the core agent, which is equipped with several specialized mechanisms, Event Scraper, Timeline Updater, and Supervisor, to dynamically refine the timeline memory to obtain the final timeline.}
  \label{fig:framework}
  \Description{The Framework of TimelineReasoner.}
\end{figure*}

\section{Related Work}
\subsection{Timeline Summarization}

Timeline Summarization (TLS) aims to construct a coherent, chronologically ordered account of event evolution from a collection of documents~\cite{Allan2001TemporalSO, Chen2019LearningTA, Yu2021MultiTimeLineS, gholipour-ghalandari-ifrim-2020-examining}. The task requires modeling temporal dependencies, identifying salient events, and reducing redundancy across heterogeneous sources. Depending on whether the document collection is fixed or dynamically retrieved, TLS is commonly categorized into closed-domain and open-domain settings~\cite{Wu2025UnfoldingTH}. 

Closed-domain TLS is often framed as a specialized form of multi-document summarization, where timelines are generated from a static document set~\cite{Chieu2004QueryBE, Li2013EvolutionaryHD,TimelineGenSocial,nguyen-etal-2014-ranking}. Prior work has mainly focused on detecting salient dates via temporal features and graphical models~\cite{Chen2023FollowTT, steen-markert-2019-abstractive, gholipour-ghalandari-ifrim-2020-examining}, or identifying major events through graph-based and clustering techniques~\cite{Duan2020ComparativeTS, Yu2021MultiTimeLineS, gholipour-ghalandari-ifrim-2020-examining, you-etal-2022-joint, li-etal-2021-timeline}. While effective in constrained scenarios, these approaches typically rely on heuristic temporal signals and struggle to capture long-range event dependencies and global narrative coherence.

More recently, LLMs have influenced TLS research, particularly in open-domain settings where document collections are dynamically retrieved. Hu et al.~\cite{hu-etal-2024-moments} leveraged LLMs to generate event-centric summaries and cluster them into timelines, while Wu et al.~\cite{Wu2025UnfoldingTH}  proposed an iterative self-questioning framework to guide retrieval and refinement. Although these methods improve flexibility, they largely treat LLMs as generative or prompt-driven components, without explicitly modeling timeline construction as a structured reasoning process. In contrast, our work explores how LRMs can support TLS through hierarchical reasoning and search, enabling dynamic evidence acquisition, iterative refinement, and principled temporal consistency verification.

\subsection{Large Reasoning Model (LRM)}

LRMs are distinguished from conventional LLMs by their emphasis on improving test-time performance through extended reasoning processes, rather than solely scaling parameters or data~\cite{Chen2025TowardsRE, Henighan2020ScalingLF, Li2025FromS1, Yang2024Qwen25TR}. Unlike traditional LLMs that operate primarily as direct text generators, LRMs incorporate structured reasoning workflows that iteratively refine intermediate representations. Representative models such as OpenAI-o1~\cite{Contributors2024OpenAIOS}, Qwen-QwQ~\cite{qwq32b}, and DeepSeek-R1~\cite{DeepSeekAI2025DeepSeekR1IR} employ explicit Chain-of-Thought (CoT) reasoning~\cite{Wei2022ChainOT}, achieving strong performance in complex tasks including mathematical reasoning and programming~\cite{Li2025FromS1}.

Existing studies identify three main directions for developing o1-level reasoning capabilities~\cite{DeepSeekAI2025DeepSeekR1IR, Chen2025AnES, Ferrag2025ReasoningBL, Hu2025OpenReasonerZeroAO, Tang2025UnlockingGL}: exposure to flawed reasoning patterns during training~\cite{Ye2024PhysicsOL}; curriculum-based enhancement using distilled reasoning data~\cite{Min2024ImitateEA}; and reinforcement learning for extended CoT capabilities~\cite{DeepSeekAI2025DeepSeekR1IR, Chen2025AnES, Hu2025OpenReasonerZeroAO}. Despite these advances, LRMs remain constrained by fixed parametric knowledge, limiting their ability to incorporate time-sensitive external information.

From a task perspective, the structured reasoning capabilities of LRMs align naturally with the requirements of TLS, which demands multi-step temporal inference, cross-document abstraction, and the ability to revise intermediate representations. Unlike prior agent-based or prompt-driven LLM approaches, which typically rely on fixed retrieval patterns or single-pass reasoning, LRMs maintain persistent reasoning states that can be incrementally updated with newly retrieved evidence. This capability is crucial for constructing timelines that are both temporally consistent and complete, as it allows the model to iteratively detect missing events, revise earlier conclusions, and integrate evidence across documents. In this way, LRMs provide a principled foundation for TimelineReasoner’s reasoning-driven approach to open-domain TLS.

\subsection{Deep Research}

Deep Research refers to reasoning-centric frameworks that enable LLMs to autonomously conduct multi-step information acquisition and evidence-based synthesis over open and evolving knowledge sources. Unlike conventional retrieval-augmented generation, which mainly focuses on augmenting generation with external documents, Deep Research frameworks emphasize iterative reasoning loops that tightly interleave planning, targeted retrieval, verification, and memory updates~\cite{Nakano2021WebGPTBQ, Komeili2021InternetAugmentedDG, Li2025Searcho1AS}.

Early examples such as WebGPT~\cite{Nakano2021WebGPTBQ} decompose complex questions into search queries and synthesize answers from retrieved evidence. Recent agent-based frameworks like WebThinker~\cite{webthinker} and WebWeaver~\cite{webwaver}, integrate reasoning with structured retrieval, combining planning, evidence acquisition, and synthesis. Tongyi DeepResearch~\cite{tongyideepresearch} further introduces an agentic LLM trained end‑to‑end with synthetic interaction data and reinforcement learning to support deep, multi‑step information‑seeking tasks.

Despite these advances, existing Deep Research systems focus on open-domain question-answering or report generation, and rarely target timeline-centric reasoning. Temporal dependencies are typically implicit, lacking mechanisms for detecting missing events or ensuring chronological consistency. In contrast, TimelineReasoner frames TLS as a Deep Research problem. This enables dynamic evidence acquisition, iterative refinement, and principled temporal reasoning, systematically addressing missing information and maintaining coherent event evolution in evolving news corpora.

\section{Methodology}

\subsection{Overview}

TimelineReasoner is a reasoning-driven framework for timeline summarization that treats timeline construction as an iterative reasoning and revision process rather than single-pass generation. It adopts a coarse-to-fine design that separates \textit{global event comprehension} from \textit{gap-driven detail refinement}, enabling the model to explicitly detect missing events and temporal inconsistencies. As illustrated in Figure~\ref{fig:framework}, TimelineReasoner consists of two main stages—\textbf{Global Cognition} and \textbf{Detailed Exploration}—supported by specialized mechanisms for structured event acquisition, incremental timeline updating, and meta-level supervision, with the LRM serving as the central controller for reasoning and retrieval.

\subsection{Global Cognition}

When constructing a timeline, human analysts often first form a coarse-grained but comprehensive understanding of how events unfold over time before diving into the details. Inspired by this process, TimelineReasoner incorporates a \textbf{Global Cognition} stage, in which the LRM synthesizes a temporally organized summary of the evolving event landscape. This high-level overview serves as a scaffold for subsequent, more fine-grained reasoning, allowing the system to anticipate gaps, track major developments, and maintain temporal coherence from the outset.
 
The main output of the Global Cognition stage is the \textbf{global event memory} $\xi$, a structured set of temporally anchored events. Each event consists of a timestamp, a concise description, and associated entities. Importantly, $\xi$ is not intended to be the final timeline; rather, it provides a stable, high-level representation that guides downstream reasoning in the Detailed Exploration stage. By explicitly separating global structure from local detail, this design ensures that TimelineReasoner can reason about coverage, event ordering, and missing information in a principled manner.
 
\subsubsection{Initializing and Updating Global Memory}
To construct this global event memory, TimelineReasoner combines structured evidence acquisition with reasoning in a coherent loop. Given an input query $q$, the \textbf{Event Scraper} retrieves relevant news documents and extracts lightweight \textbf{event metadata} $ m$, which includes temporal markers and short descriptions. To manage large collections efficiently, articles are divided into semantically coherent chunks, each processed independently. The resulting metadata fragments are then synthesized by the LRM to produce an initial global event memory:
\begin{align*}
    \xi^{(0)} = LRM\left(I, m\right),m=\textsc{EventScraper}(q),
\end{align*}
where $I$ is the instruction prompt guiding structured reasoning.

Crucially, the global event memory is designed to be dynamic rather than static. As new evidence is discovered during subsequent iterations, the memory is incrementally updated:
\begin{align*}
    \xi^{(t)} = LRM\left(\xi^{(t-1)}, m^{(t)}\right),
\end{align*}
where $m^{(t)}$ represents newly retrieved event metadata. This iterative update mechanism allows the global memory to remain temporally consistent, reflect newly acquired information, and provide a continuously reliable scaffold for the reasoning process. By explicitly modeling a dynamic, high-level event representation, TimelineReasoner mirrors human analysts’ strategy of maintaining a constantly updated mental model of the evolving situation, ensuring that subsequent detailed exploration is guided by a coherent global perspective.

\begin{table}[t]
\caption{Illustrative example of event metadata and global event memory.}
\label{tab:example_prompt}
\small
\begin{tabular}{p{0.95\linewidth}}
\toprule
\textbf{News Query}: ``Apple's pivotal product announcements'' \\ \midrule
\textbf{Extracted Event Metadata (Excerpt):} \\
\textit{January 24, 1984:} Unveiling of the Macintosh computer. \\
\textit{November 10, 2001:} Introduction of the iPod. \\ \midrule
\textbf{Global Event Memory $\xi$ (Abbreviated):} \\
January 24, 1984: Unveiling of the Macintosh Computer. This event introduced the first mass-market personal computer...November 10, 2001: Introduction of the iPod. This announcement marked Apple's successful entry into the consumer electronics market....\\ 
\bottomrule
\end{tabular}
\end{table}

\subsection{Detailed Exploration}

\subsubsection{Gap-Driven Refinement}
While Global Cognition provides a coarse-grained overview of the event landscape, it inevitably abstracts away fine-grained details and may miss less salient but temporally important events. To progressively refine the timeline, TimelineReasoner enters the \textbf{Detailed Exploration} stage, where the focus shifts from global structure to resolving specific informational deficiencies.

In this stage, TimelineReasoner main agent jointly reasons over the global event memory $\xi$ and the current timeline memory $\mathcal{M}_{\mathcal{T}}$ to identify explicit gaps, including missing events, ambiguous or coarse timestamps, and under-specified event descriptions. Rather than exhaustively retrieving additional documents—which would introduce redundancy and noise—the framework adopts a \textbf{gap-driven strategy}, where each iteration concentrates on a small number of well-defined deficiencies. This design keeps the reasoning process targeted and allows retrieval to be directly guided by the evolving timeline state.

Based on the identified gaps, the agent formulates targeted search queries $sq^{(k)}$, which are executed by the Event Scraper to retrieve refined and temporally relevant evidence. In this way, retrieval is tightly coupled with reasoning needs, ensuring that newly acquired information directly contributes to improving timeline completeness and temporal precision.

\subsubsection{Iterative Timeline Refinement}
Once sufficient evidence is obtained for the identified gaps, the agent synthesizes the retrieved information into a localized sub-timeline $\mathcal{T}^{(k)}$, capturing newly clarified or previously missing events within a limited temporal scope. Constructing sub-timelines, rather than immediately revising the entire timeline, allows TimelineReasoner to isolate local updates and reduce the risk of propagating errors across the timeline.

The generated sub-timeline is then incrementally integrated into the existing timeline memory via the \textbf{Timeline Updater}, which reconciles new evidence with previously established events and enforces consistency with the global event memory:
\begin{align*}
    \mathcal{M}_{\mathcal{T}}^{(k)} = \textsc{TimelineUpdater}\left(\mathcal{M}_{\mathcal{T}}^{(k-1)}, \mathcal{T}^{(k)}, \xi\right).
\end{align*}
This integration mechanism enables the timeline to evolve gradually through revision and refinement, rather than committing to a fixed structure in a single pass.

After each update, the \textbf{Supervisor} performs a meta-level evaluation of the current timeline memory, assessing semantic completeness, event coverage, and temporal density. This step is necessary because long reasoning chains and accumulated updates may obscure unresolved gaps or introduce subtle inconsistencies. If deficiencies remain, the Supervisor proposes a structured search plan to guide the next iteration of gap-driven reasoning and retrieval; otherwise, the process terminates. Through this iterative reasoning–retrieval–integration loop, TimelineReasoner converges to a timeline that is both comprehensive and temporally coherent.

\subsection{Specialized Mechanisms}

TimelineReasoner relies on several specialized mechanisms that are integral to its reasoning-driven framework. These mechanisms, including \textbf{Event Scraper}, \textbf{Timeline Updater}, and \textbf{Supervisor}, enable the LRM to iteratively acquire evidence, refine timeline memory, and enforce temporal consistency. These modules implement the core operational capabilities required for coherent and complete timeline construction, including retrieval, integration, and quality control tasks.

\subsubsection{Event Scraper}
The Event Scraper is responsible for structured evidence acquisition from unstructured documents. News articles $A=[a_1,a_2,\dots,a_n]$ are divided into manageable chunks $c_i$ of size $m$ to balance efficiency and information coverage. For each chunk, the model extracts \textbf{event metadata} including timestamps and concise descriptions:
\begin{align*}
    {event\_metadata}_i = LLM\{I, c_i\},
\end{align*}
where $LLM$ denotes language model inference and $I$ is the instruction prompt. Aggregating all chunks produces a comprehensive event metadata set that forms the basis for global event reasoning. This mechanism ensures that important events are not overlooked and provides the LRM with structured input for downstream reasoning, addressing the common TLS challenge of incomplete or fragmented event coverage.

\subsubsection{Timeline Updater} 
The Timeline Updater implements incremental memory integration, allowing the evolving timeline to incorporate newly retrieved events while maintaining consistency with the global event memory $\xi$. Given the current timeline memory $\mathcal{M}_\mathcal{T}$ and a newly generated sub-timeline $\mathcal{T}$, the LRM produces:
\begin{align*}
    \mathcal{M}_\mathcal{T} = LRM\left(I, \mathcal{M}_\mathcal{T}, \xi, \mathcal{T}\right).
\end{align*}
This update process ensures that previously established events are preserved or refined as new evidence becomes available. By continuously revising timeline memory, the system can correct temporal inconsistencies and gradually build a coherent global timeline, addressing the TLS problem of static, single-pass generation.

\subsubsection{Supervisor} 
The Supervisor performs meta-level reasoning over the evolving timeline memory, identifying gaps or inconsistencies that the main reasoning model may miss due to long reasoning chains or complex dependencies. It evaluates $\mathcal{M}_\mathcal{T}$ along three dimensions:
\begin{itemize}
    \item \textbf{Semantic Completeness}: Ensures that events are described in sufficient detail; if not, generates a search plan to retrieve additional information.
    \item \textbf{Coverage Consistency}: Compares the global event memory and current timeline to detect missing events, triggering targeted retrieval when necessary.
    \item \textbf{Temporal Density}: Detects sparsity or event clustering in the timeline, guiding retrieval to fill temporal gaps.
\end{itemize}

Formally, the Supervisor generates a search plan as:
\begin{align*}
    plan = LRM\left(I, \mathcal{M}_\mathcal{T}\right).
\end{align*}
This plan guides the reasoning model in subsequent iterations, allowing TimelineReasoner to actively detect missing events, refine existing entries, and enforce chronological consistency.

\section{Experiment}

\begin{table*}
\centering
\caption{Experimental results on Open-TLS. The boldfaced score in each column denotes the best result. And the \textit{underlined} score marks the second-highest result among the compared methods. Relative improvements (\%) of TimelineReasoner over the best baseline are reported in parentheses.}
\small
\scalebox{0.98}{
\begin{tabular}{llcccccccc}
\toprule
\multirow{2}{*}{ \textbf{Method}}
& \multirow{2}{*}{ \textbf{Model}} & \multicolumn{2}{c}{\textbf{Concat F1}} & \multicolumn{2}{c}{\textbf{Agree F1}} & \multicolumn{2}{c
}{\textbf{Align F1}} & \textbf{Date F1} \\
\cmidrule(lr){3-4} \cmidrule(lr){5-6} \cmidrule(lr){7-8}
 &  & ROUGE-1 & ROUGE-2 & ROUGE-1 & ROUGE-2 & ROUGE-1 & ROUGE-2 &  \\
\midrule
\multirow{3}{*}{\textbf{DIRECT}} 
&  Qwen2.5-32B& 0.2611 & 0.0587 & 0.0382 & 0.0131 & 0.0408 & 0.0138 & 0.1473\\
&  QwQ-32B&0.2682 & 0.0554 & 0.0457 & 0.0142 & 0.0487 & 0.0152 & 0.1824\\
& Qwen2.5-72B & 0.2908 & 0.0673 & 0.0559 & 0.0188 & 0.0613 & 0.0210 & 0.2102 \\
\midrule
\multirow{3}{*}{\textbf{REWRITE}}
&  Qwen2.5-32B& 0.2779 & 0.0620 & 0.0413 & 0.0122 & 0.0447 & 0.0142 & 0.1631\\
&  QwQ-32B& 0.2862 & 0.0609 & 0.0527 & 0.0170 & 0.0580 & 0.0179 & 0.1892\\
 & Qwen2.5-72B &  0.2894 & 0.0740 & 0.0608 & 0.0302 & 0.0649 & 0.0308 & 0.1993\\
\midrule
\multirow{3}{*}{\textbf{ITER\_RAG}}
&  Qwen2.5-32B& 0.3040 & 0.0702 & 0.0499 & 0.0180 & 0.0568 & 0.0188 & 0.1901\\
&  QwQ-32B& 0.2919 & 0.0639 & 0.0522 & 0.0183 & 0.0570 & 0.0184 & 0.1893\\
 & Qwen2.5-72B & 0.3292 & 0.0762 & 0.0579 & 0.0201 & 0.0633 & 0.0211 & 0.1902\\
\midrule
\multirow{3}{*}{\textbf{CHRONOS}} 
 &  Qwen2.5-32B& 0.3121 & 0.0760 & 0.0893 & \underline{0.0311} & 0.0909 & 0.0301 & 0.3119\\
 & QwQ-32B & 0.3219 & 0.0752 & 0.0901 & 0.0292 & 0.0931 & 0.0292 & 0.3030\\
  & Qwen2.5-72B & \underline{0.3380} & \underline{0.0801} & \underline{0.0962} & 0.0302 & \underline{0.0991} & \underline{0.0332} & \underline{0.3320} \\
 \midrule
 \multirow{1}{*}{\textbf{TimelineReasoner}} 
    & QwQ-32B   & \textbf{0.3631} {\scriptsize (+7.4\%)}  & \textbf{0.0894} {\scriptsize (+11.6\%)} 
& \textbf{0.1130} {\scriptsize (+17.5\%)} 
& \textbf{0.0400} {\scriptsize (+28.6\%)} 
& \textbf{0.1233} {\scriptsize (+24.4\%)} 
& \textbf{0.0411} {\scriptsize (+23.8\%)} 
& \textbf{0.3765} {\scriptsize (+13.4\%)} \\
\bottomrule
\end{tabular}}
\label{table1}
\end{table*}
\begin{table*}[ht]
\centering
\caption{Closed-domain TLS performance comparison. The boldfaced score indicates the highest result, while the \textit{underlined} score marks the second-highest result among the compared methods. Relative improvements (\%) of TimelineReasoner over the best baseline are reported in parentheses.}
\small
\label{tab:closed-domain}

\begin{tabular}{llccccccc}
\toprule
\multirow{2}{*}{ \textbf{Method}}
& \multirow{2}{*}{ \textbf{Model}} & \multicolumn{3}{c}{\textbf{Crisis}} & \multicolumn{3}{c}{\textbf{T17}} \\
\cmidrule(lr){3-5} \cmidrule(lr){6-8}
&  & \textbf{Align R-1} & \textbf{Align R-2} & \textbf{Date F1} & \textbf{Align R-1} &\textbf{ Align R-2 }& \textbf{Date F1} \\
\midrule
\textbf{CLUST}& & 0.0610 & 0.0131 & 0.2260 & 0.0822 & 0.0201 & 0.4070 \\
\midrule
\textbf{EGC} &  & 0.0791 & 0.0149 & \underline{0.2911} & \textbf{0.1031} & 0.0239 & \textbf{0.5501} \\
\midrule
\multirow{2}{*}{\textbf{LLM-TLS}}
& Qwen2.5-32B & 0.0701 & 0.0199 & 0.2732 & 0.0878 & 0.0204 & 0.5157 \\
& QwQ-32B &  0.0678 & 0.0189 & 0.2700 & 0.0852 & 0.0178 & 0.5133 \\
\midrule
\multirow{2}{*}{\textbf{DIRECT}} 
&  Qwen2.5-32B  & 0.0162 & 0.0040 & 0.0639 & 0.0537 & 0.0132 & 0.3408\\
&  QwQ-32B & 0.0184 & 0.0020 & 0.0678 & 0.0510 & 0.0140 & 0.3064\\
\midrule
\multirow{2}{*}{\textbf{REWRITE}}
&  Qwen2.5-32B  & 0.0342 & 0.0190 & 0.0927 & 0.0568 & 0.0160 & 0.3619\\
&  QwQ-32B  &  0.0441 & 0.0243 & 0.1030 &  0.0552 & 0.0139 & 0.3492\\
\midrule
\multirow{2}{*}{\textbf{ITER\_RAG}}
&  Qwen2.5-32B & 0.0552 & 0.0130 & 0.1279 & 0.0682 & 0.0201   & 0.3810\\
&  QwQ-32B & 0.0579 & 0.0173 & 0.2201 & 0.0848 & \underline{0.0282} & 0.4008\\
\midrule
\multirow{2}{*}{\textbf{CHRONOS}} 
 &  Qwen2.5-32B  & \underline{0.0861}  & \underline{0.0295} & 0.2583 & 0.0811 & 0.0220 & 0.3822\\
 & QwQ-32B  & 0.0828 & 0.0256 & 0.2907 & 0.0901 & 0.0234 & 0.4730\\
 \midrule
 \multirow{1}{*}{\textbf{TimelineReasoner}} 
& QwQ-32B 
& \textbf{0.1080} {\scriptsize (+25.4\%)} 
& \textbf{0.0331} {\scriptsize (+12.2\%)} 
& \textbf{0.3509} {\scriptsize (+20.5\%)}  
& \underline{0.1012} {\scriptsize (-1.8\%)} 
& \textbf{0.0322} {\scriptsize (+14.2\%)} 
& \underline{0.5453} {\scriptsize (-0.9\%)} \\
\bottomrule
\end{tabular}
\end{table*}
\begin{table*}[ht]
\centering
\caption{Impact of retrieval scale ($N$) on Open-TLS performance. We evaluate the impact of the number of documents in the Initialization of Global Event Memory ($N_{init}$) and the Detailed Exploration ($N_{exp}$). Bold indicates the best, \textit{underlined} indicates the second best.}
\label{tab:parameter_sensitivity}
\begin{tabular}{llcccccccc}
\toprule
\multirow{2}{*}{\textbf{Parameter}} & \multirow{2}{*}{$N$} & \multicolumn{2}{c}{\textbf{Concat F1}} & \multicolumn{2}{c}{\textbf{Agree F1}} & \multicolumn{2}{c
}{\textbf{Align F1}} & \textbf{Date F1} \\
\cmidrule(lr){3-4} \cmidrule(lr){5-6} \cmidrule(lr){7-8}
 &  & ROUGE-1 & ROUGE-2 & ROUGE-1 & ROUGE-2 & ROUGE-1 & ROUGE-2 &  \\
\midrule
 \multirow{4}{*}{\textbf{$N_{init}$}} 
  & 10  & 0.3333 & 0.0739 & 0.0811 & 0.0260 & 0.0926 & 0.0331 & 0.3336 \\
    &  20 & \textbf{0.3631} & \textbf{0.0884} & \textbf{0.1130} & \textbf{0.0400} & \textbf{0.1233} & \textbf{0.0411} & \underline{0.3764}\\
  & 30  & 0.3357 & 0.0768 & 0.0966 & 0.0322 & 0.1035 & 0.0358 & \textbf{0.3777} \\
    &  40 & \underline{0.3434} & \underline{0.0821} &\underline{ 0.0992} & \underline{0.0346} & \underline{0.1068} & \underline{0.0362} & 0.3535 \\
    \midrule
     \multirow{4}{*}{$N_{exp}$} 
    & 10  & \underline{0.3361} & 0.0759 & 0.0933 & 0.0277 & 0.1050 & 0.0298  & 0.3352 \\
    &  20 & \textbf{0.3631} & \textbf{0.0884} & \textbf{0.1130} & \textbf{0.0400} & \textbf{0.1233} & \textbf{0.0411} & \textbf{0.3764}\\
  & 30  & 0.3357 & \underline{0.0792} & \underline{0.0949} & \underline{0.0309} & \underline{0.1058} & \underline{0.0328} & \underline{0.3534}\\
    &  40 & 0.3134 & 0.0699 & 0.0859 & 0.0264 & 0.0969 & 0.0285 & 0.3513 \\
\bottomrule
\end{tabular}
\end{table*}
\subsection{Implementation Details }

To ensure reproducibility, all reported results are averaged over 3 independent runs with different random seeds. Experiments are designed to maintain stable outputs, and key hyperparameters are explicitly specified below.

\subsubsection{Model Usage}

For our method, we adopt the open-source LLM \textbf{QwQ-32B} as the main reasoning model due to its strong multi-step reasoning capabilities. The Event Scraper uses \textbf{Qwen2.5-32B} to ensure architectural consistency. Key hyperparameters for QwQ-32B are: maximum sequence length 32,768 tokens, temperature 0.7, top-$p$ 0.9, and repetition penalty 1.05. For Qwen2.5-Instruct, the maximum sequence length is 8,192 tokens, while all other hyperparameters remain identical. 

\subsubsection{Search Engine} For open-domain TLS datasets, relevant documents are retrieved via Google Serper API\footnote{https://serper.dev/}. Retrieved web pages are processed using JINA Reader\footnote{https://jina.ai/reader/} and converted into Markdown format for structured parsing. For closed-domain datasets such as crisis and t17, we construct an indexed news article collection using Elasticsearch~\cite{Gormley2015ElasticsearchTD}. Retrieval mimics realistic search behavior within the closed corpus. In both types of datasets, we retrieve the top\_$k$ ($k$$=$20) documents for each query.

\subsection{Evaluation Metrics}

We evaluate generated timelines against reference timelines using the Tilse framework~\cite{martschat-markert-2017-improving,martschat-markert-2018-temporally}, which is specifically designed for TLS.
 
\subsubsection{ROUGE-N} 
We report ROUGE-1 and ROUGE-2 F1 scores to measure lexical overlap between generated and reference timelines. Following Tilse, we consider three variants:
(1) \textit{Concat F1} computes ROUGE by concatenating all date summaries to assess overall content overlap;
(2) \textit{Agree F1} calculates ROUGE only for summaries of matching dates, focusing on precision; and
(3) \textit{Align F1} aligns predicted and reference summaries based on temporal and semantic distance, then computes ROUGE with penalties for distant alignments. 
 
\subsubsection{Date F1-score (Date-F1)} Date-F1 measures the accuracy of date prediction by computing the F1 score between generated and reference event dates, evaluating temporal coverage and alignment.

\subsection{Open-domain TLS}

\subsubsection{Baseline}

We compare TimelineReasoner with four representative baselines for open-domain TLS. All methods retrieve the same number of documents to ensure fair comparison. (1) DIRECT: retrieves documents using the original query and generates the timeline in a single pass. (2) REWRITE: expands the original query into 2-3 rewritten variants, aggregates retrieved documents, and generates the timeline. (3) ITER\_RAG: performs iterative retrieval over five rounds, where each round refines the query based on the current timeline and incrementally updates the output. (4) CHRONOS~\cite{Wu2025UnfoldingTH}: an open-domain TLS framework based on iterative self-questioning to collect multi-perspective event information. For CHRONOS, we replace the original search module with Google Serper to unify the retrieval interface across methods, while keeping other settings unchanged. All baselines are evaluated using Qwen2.5-72B, Qwen2.5-32B, and QwQ-32B to control for backbone effects.

\subsubsection{Result Analysis}
As shown in Table~\ref{table1}, TimelineReasoner consistently outperforms all baselines across all evaluation metrics. Methods incorporating query rewriting or iterative retrieval substantially outperform direct generation, confirming the importance of multi-step retrieval in open-domain TLS. 
Among non-proposed baselines, Qwen2.5-72B generally achieves the strongest performance, reflecting the benefits of larger model capacity. Notably, QwQ-32B performs on par with or better than Qwen2.5-32B despite similar parameter scales, indicating the advantage of enhanced reasoning capability for timeline construction. 
TimelineReasoner achieves statistically significant gains in both ROUGE-N and Date-F1, demonstrating improved content coverage and more accurate temporal alignment. These results highlight the effectiveness of reasoning-driven, gap-aware refinement for open-domain timeline summarization. Overall, TimelineReasoner demonstrates strong robustness and effectiveness in open-domain TLS by producing timelines that are both information-rich and temporally accurate.

\subsection{Closed-domain TLS}

\subsubsection{Baseline}
We compare our method with representative closed-domain TLS approaches, including traditional and LLM-based methods. Specifically, we include: (1) CLUST~\cite{gholipour-ghalandari-ifrim-2020-examining}, a clustering-based approach using temporal frequency-weighted event aggregation; (2) EGC~\cite{li-etal-2021-timeline}, a graph-based method that selects salient events via time-aware optimal transport; (3) LLM-TLS~\cite{hu-etal-2024-moments}, a LLM-driven framework that employs LLMs as pseudo-oracles for incremental event clustering in streaming data. In addition, we evaluate several strong LLM-based baselines originally designed for open-domain TLS, including DIRECT, REWRITE, ITER\_RAG, and CHRONOS, to assess their generalization to closed-domain settings. For fair comparison, all model-based baselines are implemented using Qwen2.5-32B and QwQ-32B, consistent with our framework.

\begin{table}
  \caption{Comparison of  token consumption and performance (Align ROUGE-2) across  TLS methods}
  \label{tab:token_compare}
  \resizebox{\columnwidth}{!}{\begin{tabular}{lccc}
  \toprule
  \textbf{Method}& \textbf{CHRONOS}& \textbf{TimelineReasoner}& \textbf{LLM-TLS}  \\
 \midrule
Token & \textasciitilde 0.7M& \textasciitilde 0.9M& \textasciitilde 47M  \\
 Performance& 0.0295& \textbf{0.0331}&0.0199\\
 \bottomrule
\end{tabular}}
\end{table}

\begin{table*}[ht]
\centering
\caption{Ablation study conducted on Open-TLS. The boldfaced score indicates the highest result.}
\label{tab:ablation}
\scalebox{1.0}{
\begin{tabular}{lccccccc}
\toprule
\multirow{2}{*}{\textbf{Method}} & \multicolumn{2}{c}{\textbf{Concat F1}} & \multicolumn{2}{c}{\textbf{Agree F1}} & \multicolumn{2}{c
}{\textbf{Align F1}} & \textbf{Date F1} \\
\cmidrule(lr){2-3} \cmidrule(lr){4-5} \cmidrule(lr){6-7}
&  ROUGE-1 & ROUGE-2 & ROUGE-1 & ROUGE-2 & ROUGE-1 & ROUGE-2 &  \\
\midrule
\textbf{Ours} & \textbf{0.363} & \textbf{0.089} & \textbf{0.113} & \textbf{0.040} & \textbf{0.123} & \textbf{0.041} & \textbf{0.376}\\
\midrule
\quad  w/o Update Global Event Memory & 0.330 & 0.073 & 0.088 & 0.027 & 0.095 & 0.032 & 0.338\\
\quad  w/o Global Event Memory& 0.298 & 0.055 & 0.038 & 0.011 & 0.045 & 0.012 & 0.154\\
\quad  w/o Timeline Updater  & 0.318 & 0.075 & 0.088 & 0.028 & 0.094 & 0.030 & 0.339\\ 
\quad    w/o Supervisor  & 0.338 & 0.082 & 0.102 & 0.036 & 0.111 & 0.037 & 0.360 \\
    \bottomrule
\end{tabular}}
\end{table*}

\subsubsection{Result Analysis}
We evaluate our method on two widely used benchmarks, \textit{Crisis} and \textit{T17}, using Aligned F1 (short for AR-1/AR-2) and Date F1 as evaluation metrics. Baseline configurations are aligned with prior work, and minor variations may arise from differences in document segmentation. The experimental results are reported in Table~\ref{tab:closed-domain}.  Overall, the results demonstrate strong cross-domain generalization from open-domain to closed-domain TLS. On the \textit{Crisis} dataset, our method outperforms all competing approaches, including CHRONOS, traditional methods, as well as the LLM-based LLM-TLS. On \textit{T17}, our approach achieves the best performance on AR-2, while ranking second on other metrics. These results indicate that our framework maintains robust performance across datasets with varying topical scopes and temporal distributions. Furthermore, the consistent improvements observed in both content-oriented metrics and temporal accuracy suggest that our method can effectively capture salient event information while preserving chronological structure. Overall, these findings confirm that our reasoning-driven framework remains effective when applied to closed-domain settings with fixed document collections.

\subsubsection{Limitations}

While TimelineReasoner achieves strong performance in closed-domain TLS, its advantages are most pronounced in settings that require iterative evidence acquisition and timeline revision. In closed-domain benchmarks with high document redundancy and clearly defined event structures, the benefits of reasoning-driven refinement may be less prominent compared to open-domain scenarios. Addressing retrieval and reasoning strategies that are better tailored to highly redundant corpora remains an interesting direction for future work.

\subsection{Cost Analysis}
\paragraph{Cost and Token Efficiency.}
To assess the cost-efficiency of TimelineReasoner, we compare the total token consumption of different TLS approaches under a closed-domain setting, where the document collection is fixed and shared across methods. This controlled setup enables a fair and reproducible comparison by eliminating variability introduced by open-domain retrieval, such as fluctuating document availability and query-dependent retrieval depth.

As shown in Table~\ref{tab:token_compare}, TimelineReasoner incurs a modest increase in token usage compared to CHRONOS, reflecting the additional reasoning steps involved in global event abstraction and coherence planning. However, it is substantially more token-efficient than LLM-TLS, which relies on long-context processing over large document collections and consequently exhibits significantly higher token consumption. These results suggest that, despite multiple LRM invocations, TimelineReasoner achieves a favorable balance between reasoning capability and computational cost by leveraging structured reasoning and selective memory construction rather than brute-force long-context generation.

\subsection{Ablation Study}

\subsubsection{Impact of Retrieval Scale}
We evaluate the sensitivity of the retrieval scale by varying the documents retrieved for Global Cognition ($N_{init}$) and Detail Exploration ($N_{exp}$) within $\{10, 20, 30, 40\}$. As shown in Table~\ref{tab:parameter_sensitivity}, $N=20$ for both stages provides the optimal balance between performance and efficiency. While $N_{init}=10$ establishes a basic event skeleton, increasing it to 20 enhances evidence diversity and ROUGE scores. However, $N_{init}=40$ leads to a performance plateau, suggesting that excessive initial data introduces redundancy that complicates the LRM's construction of global event memory. Notably, performance is more sensitive to $N_{exp}$. Insufficient retrieval ($N_{exp}=10$) hinders the Supervisor from identifying information gaps, while excessive documents ($N_{exp}>20$) introduce noise and conflicting reports, distracting the LRM during iterative reasoning. Consequently, $N=20$ serves as a robust threshold, ensuring both a coherent global perspective and high-fidelity detail refinement.

\begin{table}[t]
\footnotesize
\centering
\caption{Performance comparison under different backbone models on the Open-TLS benchmark. The best and second-best results are in bold and \textit{underlined}. DS-V3.2 is short for DeepSeek-V3.2.}
\label{tab:backbone}

\begin{tabular}{lccccccc}
\toprule
\multirow{2}{*}{\textbf{LLM}}
& \multicolumn{2}{c}{\textbf{Concat F1}} 
& \multicolumn{2}{c}{\textbf{Agree F1}} 
& \multicolumn{2}{c}{\textbf{Align F1}} 
& \textbf{Date F1} \\
\cmidrule(lr){2-3} \cmidrule(lr){4-5} \cmidrule(lr){6-7}
& R-1 & R-2 & R-1 & R-2 & R-1 & R-2 &  \\
\midrule

\multicolumn{8}{l}{\textbf{CHRONOS}} \\
QwQ-32B        & 0.322 & 0.075 & 0.090 & 0.029 & 0.093 & 0.029 & 0.303 \\
Qwen2.5-72B   & 0.338 & 0.080 & 0.096 & 0.030 & 0.099 & 0.033 & 0.332 \\
DS-V3.2       &       0.359&       \underline{0.089}&       0.097&       0.036&       0.104&       0.038&       0.351\\
\midrule

\multicolumn{8}{l}{\textbf{Ours}} \\
QwQ-32B        & \underline{0.363} & \underline{0.089} & \underline{0.113} & \underline{0.040} & \underline{0.123} & \underline{0.041} & \underline{0.376} \\
Qwen2.5-72B   & 0.326 & 0.031 & 0.077 & 0.030 & 0.083 & 0.032 & 0.240 \\
DS-V3.2       & \textbf{0.378} & \textbf{0.106} & \textbf{0.141} & \textbf{0.054} & \textbf{0.149} & \textbf{0.055} & \textbf{0.410} \\
\bottomrule
\end{tabular}
\end{table}

\subsubsection{Role of Global Event Memory}

Global event memory maintains a unified representation of event information and provides persistent guidance for both the main reasoning process and the supervisor. As shown in Table~\ref{tab:ablation}, removing global event memory leads to substantial performance degradation across all metrics, with notable drops in both Date F1 and ROUGE scores. These results indicate that global event memory is indispensable for preserving temporal accuracy and modeling inter-event relationships, serving simultaneously as a knowledge repository and a coordination mechanism for reasoning.

We further examine the impact of memory maintenance by disabling updates with event metadata after each Event Scraper call. This ablation leads to consistent but smaller performance decreases, suggesting that continuous memory refinement helps maintain an up-to-date and comprehensive event representation, which in turn supports more stable reasoning and timeline generation.

\subsubsection{Impact of Timeline Memory}

Timeline Updater serves a critical function in refining the timeline memory within our framework. When this component is removed, the system falls back to naive concatenation of subtimelines. As reported in Table~\ref{tab:ablation}, this modification leads to consistent performance degradation across all evaluation metrics, indicating reduced quality in both event description granularity and temporal coverage precision. These results confirm the updater's essential role in memory optimization, where it actively synthesizes and reorganizes temporal information rather than merely aggregating raw outputs. The observed performance drop, while moderate in scale, systematically demonstrates how Timeline Updater's hierarchical processing of temporal relationships contributes to final timeline coherence.

\subsubsection{Importance of Supervisor}

The Supervisor monitors the evolving timeline memory and identifies potential deficiencies after the main reasoning model updates the timeline memory. It then generates targeted search plans to guide subsequent retrieval, helping focus the reasoning process and avoid unnecessary computation. As shown in Table~\ref{tab:ablation}, removing the Supervisor leads to consistent performance drops across all metrics, with particularly noticeable decreases in temporal coherence and event coverage. These results suggest that the Supervisor contributes to more accurate and complete timeline construction by coordinating iterative reasoning and targeted retrieval, without which the system’s ability to detect and fill gaps is diminished.

\begin{table}
  \caption{Case study of the process of TimelineReasoner generating a timeline.}
  \label{tab:case_study}
  \begin{tabular}{p{0.45\textwidth}}
    \toprule
    \textbf{Reasoning chain of TimelineReasoner}\\
    \midrule
    You are a reasoning assistant with the ability to perform web searches to help create an accurate chronological timeline. \\
    ...\\
    The event development process:\\
    ...\\
    Now, proceed to make the necessary search queries to confirm these points and ensure accuracy.\\
    First, verifying the Mac Studio M2 Ultra's announcement date:\\
    <|begin\_search\_query|>Apple Mac Studio M2 Ultra release date<|end\_search\_query|> (Here call Event Scraper)\\
    ...\\
Let me structure each part clearly and ensure the descriptions are detailed as per the search results.\\
<|begin\_update\_timeline|> ... <|end\_update\_timeline|> (Here call Timeline Updater)\\
...\\
Wait, I think the current timeline is not comprehensive enough. I need to search for more information. Here is my future search plan: ... (Here Supervisor returns a search plan)\\
...\\
\bottomrule
\end{tabular}
\end{table}

\subsection{Backbone Robustness Analysis
}
To verify the capability of our framework to effectively exploit the reasoning capacity of large reasoning models, we evaluate multiple backbone models beyond the default QwQ-32B used in our method. Specifically, we replace the backbone with an instruction model, Qwen2.5-72B, as well as a recently released model with strong reasoning and agent capabilities, DeepSeek-V3.2. All experiments are conducted on the Open-TLS benchmark, and the results are reported in Table~\ref{tab:backbone}. 
The results show that, within our framework, the two reasoning models consistently outperform Qwen2.5-72B across multiple evaluation metrics, indicating that our method can effectively leverage enhanced reasoning abilities. Moreover, when comparing reasoning backbones, DeepSeek-V3.2 achieves clear performance gains over QwQ-32B, suggesting that stronger reasoning backbones further amplify the effectiveness of our approach. In contrast, when applied to the baseline, DeepSeek-V3.2 does not demonstrate a comparable advantage: its performance is substantially lower than that achieved by DeepSeek-V3.2 within our framework, and is roughly on par with QwQ-32B when used as the backbone of our method. These findings demonstrate that our method not only benefits from stronger reasoning backbones, but also more effectively translates the reasoning capacity of large models into improved performance for complex timeline summarization tasks.

\subsection{Case Study} 
To illustrate how TimelineReasoner performs reasoning-driven timeline construction, we present a case study on the news query \textit{Apple’s pivotal product announcements}, as shown in Table~\ref{tab:case_study}. Starting from the global event memory, the reasoning model identifies a missing temporal detail for a salient event (the announcement date of the Mac Studio M2 Ultra). To resolve this gap, it formulates a targeted query, which is executed by the Event Scraper to retrieve temporally relevant evidence. The retrieved information is synthesized into a localized sub-timeline and incrementally integrated into the existing timeline via the Timeline Updater, refining the timeline without altering previously confirmed events. After the update, the Supervisor evaluates the timeline and detects remaining coverage deficiencies, generating a structured search plan to guide further reasoning and retrieval. This iterative process continues until the timeline satisfies completeness and temporal consistency criteria. This case study highlights how TimelineReasoner explicitly detects missing information, performs gap-driven retrieval, and revises the timeline through iterative reasoning.

\section{Conclusion and Future Work }

In this work, we propose TimelineReasoner, an active, reasoning-driven framework for TLS. By leveraging the structured reasoning capabilities of LRMs, our framework decouples macroscopic event tracking from fine-grained detail refinement through a two-stage paradigm: Global Cognition and Detail Exploration. Supported by specialized mechanisms, TimelineReasoner maintains a dynamic timeline memory that is iteratively refined with targeted evidence.
Extensive evaluations demonstrate that TimelineReasoner significantly outperforms LLM-based baselines in open-domain TLS tasks and remains highly competitive in closed-domain settings. Our analysis further reveals that the framework is uniquely capable of translating the latent reasoning power of LRMs into tangible performance gains while maintaining a favorable balance between output fidelity and token efficiency.

In the future, we plan to explore other ways to enhance the performance of our method. First, we plan to apply reinforcement learning to develop specialized reward functions, optimizing the specificity of generated queries and the strategic quality of search plans. Second, we will investigate adaptive reasoning strategies tailored for redundant information environments. Finally, we aim to extend the framework’s memory scaling capabilities to support the construction of comprehensive timelines for event sequences spanning multiple years.


\bibliographystyle{ACM-Reference-Format}
\bibliography{sample-base}

@String{Computer = "{IEEE} Computer" }

@String{Chelsea = "Chelsea" }

@article{Chen2023FollowTT,
  author       = {Xiuying Chen and
                  Mingzhe Li and
                  Shen Gao and
                  Zhangming Chan and
                  Dongyan Zhao and
                  Xin Gao and
                  Xiangliang Zhang and
                  Rui Yan},
  title        = {Follow the Timeline! Generating an Abstractive and Extractive Timeline
                  Summary in Chronological Order},
  journal      = {{ACM} Trans. Inf. Syst.},
  volume       = {41},
  number       = {1},
  pages        = {9:1--9:30},
  year         = {2023},
  url          = {https://doi.org/10.1145/3517221},
  doi          = {10.1145/3517221},
  bibsource    = {dblp computer science bibliography, https://dblp.org}
}

@inproceedings{nguyen-etal-2014-ranking,
  author       = {Kiem{-}Hieu Nguyen and
                  Xavier Tannier and
                  V{\'{e}}ronique Moriceau},
  editor       = {Jan Hajic and
                  Junichi Tsujii},
  title        = {Ranking Multidocument Event Descriptions for Building Thematic Timelines},
  booktitle    = {{COLING} 2014, 25th International Conference on Computational Linguistics,
                  Proceedings of the Conference: Technical Papers, August 23-29, 2014,
                  Dublin, Ireland},
  pages        = {1208--1217},
  publisher    = {{ACL}},
  year         = {2014},
  url          = {https://aclanthology.org/C14-1114/},
  bibsource    = {dblp computer science bibliography, https://dblp.org}
}

@inproceedings{steen-markert-2019-abstractive,
    title = "Abstractive Timeline Summarization",
    author = "Steen, Julius  and
      Markert, Katja",
    editor = "Wang, Lu  and
      Cheung, Jackie Chi Kit  and
      Carenini, Giuseppe  and
      Liu, Fei",
    booktitle = "Proceedings of the 2nd Workshop on New Frontiers in Summarization",
    month = nov,
    year = "2019",
    address = "Hong Kong, China",
    publisher = "Association for Computational Linguistics",
    url = "https://aclanthology.org/D19-5403/",
    doi = "10.18653/v1/D19-5403",
    pages = "21--31",
}

@inproceedings{Allan2001TemporalSO,
  author       = {James Allan and
                  Rahul Gupta and
                  Vikas Khandelwal},
  editor       = {W. Bruce Croft and
                  David J. Harper and
                  Donald H. Kraft and
                  Justin Zobel},
  title        = {Temporal Summaries of News Topics},
  booktitle    = {{SIGIR} 2001: Proceedings of the 24th Annual International {ACM} {SIGIR}
                  Conference on Research and Development in Information Retrieval, September
                  9-13, 2001, New Orleans, Louisiana, {USA}},
  pages        = {10--18},
  publisher    = {{ACM}},
  year         = {2001},
  url          = {https://doi.org/10.1145/383952.383954},
  doi          = {10.1145/383952.383954},
  bibsource    = {dblp computer science bibliography, https://dblp.org}
}

@inproceedings{Chen2019LearningTA,
  author       = {Xiuying Chen and
                  Zhangming Chan and
                  Shen Gao and
                  Meng{-}Hsuan Yu and
                  Dongyan Zhao and
                  Rui Yan},
  editor       = {Sarit Kraus},
  title        = {Learning towards Abstractive Timeline Summarization},
  booktitle    = {Proceedings of the Twenty-Eighth International Joint Conference on
                  Artificial Intelligence, {IJCAI} 2019, Macao, China, August 10-16,
                  2019},
  pages        = {4939--4945},
  publisher    = {ijcai.org},
  year         = {2019},
  url          = {https://doi.org/10.24963/ijcai.2019/686},
  doi          = {10.24963/IJCAI.2019/686},
  bibsource    = {dblp computer science bibliography, https://dblp.org}
}

@inproceedings{Duan2020ComparativeTS,
  author       = {Yijun Duan and
                  Adam Jatowt and
                  Masatoshi Yoshikawa},
  editor       = {Giuseppe De Giacomo and
                  Alejandro Catal{\'{a}} and
                  Bistra Dilkina and
                  Michela Milano and
                  Sen{\'{e}}n Barro and
                  Alberto Bugar{\'{\i}}n and
                  J{\'{e}}r{\^{o}}me Lang},
  title        = {Comparative Timeline Summarization via Dynamic Affinity-Preserving
                  Random Walk},
  booktitle    = {{ECAI} 2020 - 24th European Conference on Artificial Intelligence,
                  29 August-8 September 2020, Santiago de Compostela, Spain, August
                  29 - September 8, 2020 - Including 10th Conference on Prestigious
                  Applications of Artificial Intelligence {(PAIS} 2020)},
  series       = {Frontiers in Artificial Intelligence and Applications},
  volume       = {325},
  pages        = {1778--1785},
  publisher    = {{IOS} Press},
  year         = {2020},
  url          = {https://doi.org/10.3233/FAIA200292},
  doi          = {10.3233/FAIA200292},
  bibsource    = {dblp computer science bibliography, https://dblp.org}
}

@inproceedings{gholipour-ghalandari-ifrim-2020-examining,
  author       = {Demian Gholipour Ghalandari and
                  Georgiana Ifrim},
  editor       = {Dan Jurafsky and
                  Joyce Chai and
                  Natalie Schluter and
                  Joel R. Tetreault},
  title        = {Examining the State-of-the-Art in News Timeline Summarization},
  booktitle    = {Proceedings of the 58th Annual Meeting of the Association for Computational
                  Linguistics, {ACL} 2020, Online, July 5-10, 2020},
  pages        = {1322--1334},
  publisher    = {Association for Computational Linguistics},
  year         = {2020},
  url          = {https://doi.org/10.18653/v1/2020.acl-main.122},
  doi          = {10.18653/V1/2020.ACL-MAIN.122},
  bibsource    = {dblp computer science bibliography, https://dblp.org}
}

@inproceedings{TimelineGenSocial,
  author       = {Wayne Xin Zhao and
                  Yanwei Guo and
                  Rui Yan and
                  Yulan He and
                  Xiaoming Li},
  editor       = {Gareth J. F. Jones and
                  Paraic Sheridan and
                  Diane Kelly and
                  Maarten de Rijke and
                  Tetsuya Sakai},
  title        = {Timeline generation with social attention},
  booktitle    = {The 36th International {ACM} {SIGIR} conference on research and development
                  in Information Retrieval, {SIGIR} '13, Dublin, Ireland - July 28 -
                  August 01, 2013},
  pages        = {1061--1064},
  publisher    = {{ACM}},
  year         = {2013},
  url          = {https://doi.org/10.1145/2484028.2484103},
  doi          = {10.1145/2484028.2484103},
  bibsource    = {dblp computer science bibliography, https://dblp.org}
}

@inproceedings{Yu2021MultiTimeLineS,
  author       = {Yi Yu and
                  Adam Jatowt and
                  Antoine Doucet and
                  Kazunari Sugiyama and
                  Masatoshi Yoshikawa},
  editor       = {Chengqing Zong and
                  Fei Xia and
                  Wenjie Li and
                  Roberto Navigli},
  title        = {Multi-TimeLine Summarization {(MTLS):} Improving Timeline Summarization
                  by Generating Multiple Summaries},
  booktitle    = {Proceedings of the 59th Annual Meeting of the Association for Computational
                  Linguistics and the 11th International Joint Conference on Natural
                  Language Processing, {ACL/IJCNLP} 2021, (Volume 1: Long Papers), Virtual
                  Event, August 1-6, 2021},
  pages        = {377--387},
  publisher    = {Association for Computational Linguistics},
  year         = {2021},
  url          = {https://doi.org/10.18653/v1/2021.acl-long.32},
  doi          = {10.18653/V1/2021.ACL-LONG.32},
  bibsource    = {dblp computer science bibliography, https://dblp.org}
}

@inproceedings{Li2013EvolutionaryHD,
  author       = {Jiwei Li and
                  Sujian Li},
  title        = {Evolutionary Hierarchical Dirichlet Process for Timeline Summarization},
  booktitle    = {Proceedings of the 51st Annual Meeting of the Association for Computational
                  Linguistics, {ACL} 2013, 4-9 August 2013, Sofia, Bulgaria, Volume
                  2: Short Papers},
  pages        = {556--560},
  publisher    = {The Association for Computer Linguistics},
  year         = {2013},
  url          = {https://aclanthology.org/P13-2099/},
  bibsource    = {dblp computer science bibliography, https://dblp.org}
}

@inproceedings{you-etal-2022-joint,
  author       = {Jingyi You and
                  Dongyuan Li and
                  Hidetaka Kamigaito and
                  Kotaro Funakoshi and
                  Manabu Okumura},
  editor       = {Marine Carpuat and
                  Marie{-}Catherine de Marneffe and
                  Iv{\'{a}}n Vladimir Meza Ru{\'{\i}}z},
  title        = {Joint Learning-based Heterogeneous Graph Attention Network for Timeline
                  Summarization},
  booktitle    = {Proceedings of the 2022 Conference of the North American Chapter of
                  the Association for Computational Linguistics: Human Language Technologies,
                  {NAACL} 2022, Seattle, WA, United States, July 10-15, 2022},
  pages        = {4091--4104},
  publisher    = {Association for Computational Linguistics},
  year         = {2022},
  url          = {https://doi.org/10.18653/v1/2022.naacl-main.301},
  doi          = {10.18653/V1/2022.NAACL-MAIN.301},
  bibsource    = {dblp computer science bibliography, https://dblp.org}
}

@inproceedings{Chieu2004QueryBE,
  author       = {Hai Leong Chieu and
                  Yoong Keok Lee},
  editor       = {Mark Sanderson and
                  Kalervo J{\"{a}}rvelin and
                  James Allan and
                  Peter Bruza},
  title        = {Query based event extraction along a timeline},
  booktitle    = {{SIGIR} 2004: Proceedings of the 27th Annual International {ACM} {SIGIR}
                  Conference on Research and Development in Information Retrieval, Sheffield,
                  UK, July 25-29, 2004},
  pages        = {425--432},
  publisher    = {{ACM}},
  year         = {2004},
  url          = {https://doi.org/10.1145/1008992.1009065},
  doi          = {10.1145/1008992.1009065},
  bibsource    = {dblp computer science bibliography, https://dblp.org}
}

@article{Chen2025TowardsRE,
  author       = {Qiguang Chen and
                  Libo Qin and
                  Jinhao Liu and
                  Dengyun Peng and
                  Jiannan Guan and
                  Peng Wang and
                  Mengkang Hu and
                  Yuhang Zhou and
                  Te Gao and
                  Wanxiang Che},
  title        = {Towards Reasoning Era: {A} Survey of Long Chain-of-Thought for Reasoning
                  Large Language Models},
  journal      = {CoRR},
  volume       = {abs/2503.09567},
  year         = {2025},
  url          = {https://doi.org/10.48550/arXiv.2503.09567},
  doi          = {10.48550/ARXIV.2503.09567},
  eprinttype    = {arXiv},
  eprint       = {2503.09567},
  bibsource    = {dblp computer science bibliography, https://dblp.org}
}

@inproceedings{Wu2025UnfoldingTH,
  author       = {Weiqi Wu and
                  Shen Huang and
                  Yong Jiang and
                  Pengjun Xie and
                  Fei Huang and
                  Hai Zhao},
  editor       = {Luis Chiruzzo and
                  Alan Ritter and
                  Lu Wang},
  title        = {Unfolding the Headline: Iterative Self-Questioning for News Retrieval
                  and Timeline Summarization},
  booktitle    = {Findings of the Association for Computational Linguistics: {NAACL}
                  2025, Albuquerque, New Mexico, USA, April 29 - May 4, 2025},
  pages        = {4385--4398},
  publisher    = {Association for Computational Linguistics},
  year         = {2025},
  url          = {https://doi.org/10.18653/v1/2025.findings-naacl.248},
  doi          = {10.18653/V1/2025.FINDINGS-NAACL.248},
  bibsource    = {dblp computer science bibliography, https://dblp.org}
}

@article{DeepSeekAI2025DeepSeekR1IR,
  author       = {DeepSeek{-}AI},
  title        = {DeepSeek-R1: Incentivizing Reasoning Capability in LLMs via Reinforcement
                  Learning},
  journal      = {CoRR},
  volume       = {abs/2501.12948},
  year         = {2025},
  url          = {https://doi.org/10.48550/arXiv.2501.12948},
  doi          = {10.48550/ARXIV.2501.12948},
  eprinttype    = {arXiv},
  eprint       = {2501.12948},
  bibsource    = {dblp computer science bibliography, https://dblp.org}
}

@inproceedings{hu-etal-2024-moments,
  author       = {Qisheng Hu and
                  Geonsik Moon and
                  Hwee Tou Ng},
  editor       = {Lun{-}Wei Ku and
                  Andre Martins and
                  Vivek Srikumar},
  title        = {From Moments to Milestones: Incremental Timeline Summarization Leveraging
                  Large Language Models},
  booktitle    = {Proceedings of the 62nd Annual Meeting of the Association for Computational
                  Linguistics (Volume 1: Long Papers), {ACL} 2024, Bangkok, Thailand,
                  August 11-16, 2024},
  pages        = {7232--7246},
  publisher    = {Association for Computational Linguistics},
  year         = {2024},
  url          = {https://doi.org/10.18653/v1/2024.acl-long.390},
  doi          = {10.18653/V1/2024.ACL-LONG.390},
  bibsource    = {dblp computer science bibliography, https://dblp.org}
}

@inproceedings{li-etal-2021-timeline,
  author       = {Manling Li and
                  Tengfei Ma and
                  Mo Yu and
                  Lingfei Wu and
                  Tian Gao and
                  Heng Ji and
                  Kathleen R. McKeown},
  editor       = {Marie{-}Francine Moens and
                  Xuanjing Huang and
                  Lucia Specia and
                  Scott Wen{-}tau Yih},
  title        = {Timeline Summarization based on Event Graph Compression via Time-Aware
                  Optimal Transport},
  booktitle    = {Proceedings of the 2021 Conference on Empirical Methods in Natural
                  Language Processing, {EMNLP} 2021, Virtual Event / Punta Cana, Dominican
                  Republic, 7-11 November, 2021},
  pages        = {6443--6456},
  publisher    = {Association for Computational Linguistics},
  year         = {2021},
  url          = {https://doi.org/10.18653/v1/2021.emnlp-main.519},
  doi          = {10.18653/V1/2021.EMNLP-MAIN.519},
  bibsource    = {dblp computer science bibliography, https://dblp.org}
}

@article{Chen2025AnES,
  author       = {Zhipeng Chen and
                  Yingqian Min and
                  Beichen Zhang and
                  Jie Chen and
                  Jinhao Jiang and
                  Daixuan Cheng and
                  Wayne Xin Zhao and
                  Zheng Liu and
                  Xu Miao and
                  Yang Lu and
                  Lei Fang and
                  Zhongyuan Wang and
                  Ji{-}Rong Wen},
  title        = {An Empirical Study on Eliciting and Improving R1-like Reasoning Models},
  journal      = {CoRR},
  volume       = {abs/2503.04548},
  year         = {2025},
  url          = {https://doi.org/10.48550/arXiv.2503.04548},
  doi          = {10.48550/ARXIV.2503.04548},
  eprinttype    = {arXiv},
  eprint       = {2503.04548},
  bibsource    = {dblp computer science bibliography, https://dblp.org}
}

@article{Li2025Searcho1AS,
  author       = {Xiaoxi Li and
                  Guanting Dong and
                  Jiajie Jin and
                  Yuyao Zhang and
                  Yujia Zhou and
                  Yutao Zhu and
                  Peitian Zhang and
                  Zhicheng Dou},
  title        = {Search-o1: Agentic Search-Enhanced Large Reasoning Models},
  journal      = {CoRR},
  volume       = {abs/2501.05366},
  year         = {2025},
  url          = {https://doi.org/10.48550/arXiv.2501.05366},
  doi          = {10.48550/ARXIV.2501.05366},
  eprinttype    = {arXiv},
  eprint       = {2501.05366},
  bibsource    = {dblp computer science bibliography, https://dblp.org}
}

@article{Ferrag2025ReasoningBL,
  author       = {Mohamed Amine Ferrag and
                  Norbert Tihanyi and
                  M{\'{e}}rouane Debbah},
  title        = {Reasoning Beyond Limits: Advances and Open Problems for LLMs},
  journal      = {CoRR},
  volume       = {abs/2503.22732},
  year         = {2025},
  url          = {https://doi.org/10.48550/arXiv.2503.22732},
  doi          = {10.48550/ARXIV.2503.22732},
  eprinttype    = {arXiv},
  eprint       = {2503.22732},
  bibsource    = {dblp computer science bibliography, https://dblp.org}
}

@article{Nakano2021WebGPTBQ,
  author       = {Reiichiro Nakano and
                  Jacob Hilton and
                  Suchir Balaji and
                  Jeff Wu and
                  Long Ouyang and
                  Christina Kim and
                  Christopher Hesse and
                  Shantanu Jain and
                  Vineet Kosaraju and
                  William Saunders and
                  Xu Jiang and
                  Karl Cobbe and
                  Tyna Eloundou and
                  Gretchen Krueger and
                  Kevin Button and
                  Matthew Knight and
                  Benjamin Chess and
                  John Schulman},
  title        = {WebGPT: Browser-assisted question-answering with human feedback},
  journal      = {CoRR},
  volume       = {abs/2112.09332},
  year         = {2021},
  url          = {https://arxiv.org/abs/2112.09332},
  eprinttype    = {arXiv},
  eprint       = {2112.09332},
  bibsource    = {dblp computer science bibliography, https://dblp.org}
}

@article{Henighan2020ScalingLF,
  author       = {Tom Henighan and
                  Jared Kaplan and
                  Mor Katz and
                  Mark Chen and
                  Christopher Hesse and
                  Jacob Jackson and
                  Heewoo Jun and
                  Tom B. Brown and
                  Prafulla Dhariwal and
                  Scott Gray and
                  Chris Hallacy and
                  Benjamin Mann and
                  Alec Radford and
                  Aditya Ramesh and
                  Nick Ryder and
                  Daniel M. Ziegler and
                  John Schulman and
                  Dario Amodei and
                  Sam McCandlish},
  title        = {Scaling Laws for Autoregressive Generative Modeling},
  journal      = {CoRR},
  volume       = {abs/2010.14701},
  year         = {2020},
  url          = {https://arxiv.org/abs/2010.14701},
  eprinttype    = {arXiv},
  eprint       = {2010.14701},
  bibsource    = {dblp computer science bibliography, https://dblp.org}
}

@inproceedings{Komeili2021InternetAugmentedDG,
  author       = {Mojtaba Komeili and
                  Kurt Shuster and
                  Jason Weston},
  editor       = {Smaranda Muresan and
                  Preslav Nakov and
                  Aline Villavicencio},
  title        = {Internet-Augmented Dialogue Generation},
  booktitle    = {Proceedings of the 60th Annual Meeting of the Association for Computational
                  Linguistics (Volume 1: Long Papers), {ACL} 2022, Dublin, Ireland,
                  May 22-27, 2022},
  pages        = {8460--8478},
  publisher    = {Association for Computational Linguistics},
  year         = {2022},
  url          = {https://doi.org/10.18653/v1/2022.acl-long.579},
  doi          = {10.18653/V1/2022.ACL-LONG.579},
  bibsource    = {dblp computer science bibliography, https://dblp.org}
}

@article{Hu2025OpenReasonerZeroAO,
  author       = {Jingcheng Hu and
                  Yinmin Zhang and
                  Qi Han and
                  Daxin Jiang and
                  Xiangyu Zhang and
                  Heung{-}Yeung Shum},
  title        = {Open-Reasoner-Zero: An Open Source Approach to Scaling Up Reinforcement
                  Learning on the Base Model},
  journal      = {CoRR},
  volume       = {abs/2503.24290},
  year         = {2025},
  url          = {https://doi.org/10.48550/arXiv.2503.24290},
  doi          = {10.48550/ARXIV.2503.24290},
  eprinttype    = {arXiv},
  eprint       = {2503.24290},
  bibsource    = {dblp computer science bibliography, https://dblp.org}
}

@article{Contributors2024OpenAIOS,
  author       = {Aaron Jaech and
                  Adam Kalai and
                  Adam Lerer and
                  Adam Richardson and
                  Ahmed El{-}Kishky and
                  Aiden Low and
                  Alec Helyar and
                  Aleksander Madry and
                  Alex Beutel and
                  Alex Carney and
                  Alex Iftimie and
                  Alex Karpenko and
                  Alex Tachard Passos and
                  Alexander Neitz and
                  Alexander Prokofiev and
                  Alexander Wei and
                  Allison Tam and
                  Ally Bennett and
                  Ananya Kumar and
                  Andre Saraiva and
                  Andrea Vallone and
                  Andrew Duberstein and
                  Andrew Kondrich and
                  Andrey Mishchenko and
                  Andy Applebaum and
                  Angela Jiang and
                  Ashvin Nair and
                  Barret Zoph and
                  Behrooz Ghorbani and
                  Ben Rossen and
                  Benjamin Sokolowsky and
                  Boaz Barak and
                  Bob McGrew and
                  Borys Minaiev and
                  Botao Hao and
                  Bowen Baker and
                  Brandon Houghton and
                  Brandon McKinzie and
                  Brydon Eastman and
                  Camillo Lugaresi and
                  Cary Bassin and
                  Cary Hudson and
                  Chak Ming Li and
                  Charles de Bourcy and
                  Chelsea Voss and
                  Chen Shen and
                  Chong Zhang and
                  Chris Koch and
                  Chris Orsinger and
                  Christopher Hesse and
                  Claudia Fischer and
                  Clive Chan and
                  Dan Roberts and
                  Daniel Kappler and
                  Daniel Levy and
                  Daniel Selsam and
                  David Dohan and
                  David Farhi and
                  David Mely and
                  David Robinson and
                  Dimitris Tsipras and
                  Doug Li and
                  Dragos Oprica and
                  Eben Freeman and
                  Eddie Zhang and
                  Edmund Wong and
                  Elizabeth Proehl and
                  Enoch Cheung and
                  Eric Mitchell and
                  Eric Wallace and
                  Erik Ritter and
                  Evan Mays and
                  Fan Wang and
                  Felipe Petroski Such and
                  Filippo Raso and
                  Florencia Leoni and
                  Foivos Tsimpourlas and
                  Francis Song and
                  Fred von Lohmann and
                  Freddie Sulit and
                  Geoff Salmon and
                  Giambattista Parascandolo and
                  Gildas Chabot and
                  Grace Zhao and
                  Greg Brockman and
                  Guillaume Leclerc and
                  Hadi Salman and
                  Haiming Bao and
                  Hao Sheng and
                  Hart Andrin and
                  Hessam Bagherinezhad and
                  Hongyu Ren and
                  Hunter Lightman and
                  Hyung Won Chung and
                  Ian Kivlichan and
                  Ian O'Connell and
                  Ian Osband and
                  Ignasi Clavera Gilaberte and
                  Ilge Akkaya},
  title        = {OpenAI o1 System Card},
  journal      = {CoRR},
  volume       = {abs/2412.16720},
  year         = {2024},
  url          = {https://doi.org/10.48550/arXiv.2412.16720},
  doi          = {10.48550/ARXIV.2412.16720},
  eprinttype    = {arXiv},
  eprint       = {2412.16720},
  bibsource    = {dblp computer science bibliography, https://dblp.org}
}

@article{Li2025FromS1,
  author       = {Zhong{-}Zhi Li and
                  Duzhen Zhang and
                  Ming{-}Liang Zhang and
                  Jiaxin Zhang and
                  Zengyan Liu and
                  Yuxuan Yao and
                  Haotian Xu and
                  Junhao Zheng and
                  Pei{-}Jie Wang and
                  Xiuyi Chen and
                  Yingying Zhang and
                  Fei Yin and
                  Jiahua Dong and
                  Zhijiang Guo and
                  Le Song and
                  Cheng{-}Lin Liu},
  title        = {From System 1 to System 2: {A} Survey of Reasoning Large Language
                  Models},
  journal      = {CoRR},
  volume       = {abs/2502.17419},
  year         = {2025},
  url          = {https://doi.org/10.48550/arXiv.2502.17419},
  doi          = {10.48550/ARXIV.2502.17419},
  eprinttype    = {arXiv},
  eprint       = {2502.17419},
  bibsource    = {dblp computer science bibliography, https://dblp.org}
}

@article{Min2024ImitateEA,
  author       = {Yingqian Min and
                  Zhipeng Chen and
                  Jinhao Jiang and
                  Jie Chen and
                  Jia Deng and
                  Yiwen Hu and
                  Yiru Tang and
                  Jiapeng Wang and
                  Xiaoxue Cheng and
                  Huatong Song and
                  Wayne Xin Zhao and
                  Zheng Liu and
                  Zhongyuan Wang and
                  Ji{-}Rong Wen},
  title        = {Imitate, Explore, and Self-Improve: {A} Reproduction Report on Slow-thinking
                  Reasoning Systems},
  journal      = {CoRR},
  volume       = {abs/2412.09413},
  year         = {2024},
  url          = {https://doi.org/10.48550/arXiv.2412.09413},
  doi          = {10.48550/ARXIV.2412.09413},
  eprinttype    = {arXiv},
  eprint       = {2412.09413},
  bibsource    = {dblp computer science bibliography, https://dblp.org}
}

@article{Achiam2023GPT4TR,
  author       = {OpenAI},
  title        = {{GPT-4} Technical Report},
  journal      = {CoRR},
  volume       = {abs/2303.08774},
  year         = {2023},
  url          = {https://doi.org/10.48550/arXiv.2303.08774},
  doi          = {10.48550/ARXIV.2303.08774},
  eprinttype    = {arXiv},
  eprint       = {2303.08774},
  bibsource    = {dblp computer science bibliography, https://dblp.org}
}

@inproceedings{Gormley2015ElasticsearchTD,
  title={Elasticsearch: The Definitive Guide},
  author={Clinton Gormley and Zachary J. Tong},
  year={2015},
  url={https://api.semanticscholar.org/CorpusID:62964734}
}

@article{Yang2024Qwen25TR,
  author       = {An Yang and
                  Baosong Yang and
                  Beichen Zhang and
                  Binyuan Hui and
                  Bo Zheng and
                  Bowen Yu and
                  Chengyuan Li and
                  Dayiheng Liu and
                  Fei Huang and
                  Haoran Wei and
                  Huan Lin and
                  Jian Yang and
                  Jianhong Tu and
                  Jianwei Zhang and
                  Jianxin Yang and
                  Jiaxi Yang and
                  Jingren Zhou and
                  Junyang Lin and
                  Kai Dang and
                  Keming Lu and
                  Keqin Bao and
                  Kexin Yang and
                  Le Yu and
                  Mei Li and
                  Mingfeng Xue and
                  Pei Zhang and
                  Qin Zhu and
                  Rui Men and
                  Runji Lin and
                  Tianhao Li and
                  Tingyu Xia and
                  Xingzhang Ren and
                  Xuancheng Ren and
                  Yang Fan and
                  Yang Su and
                  Yichang Zhang and
                  Yu Wan and
                  Yuqiong Liu and
                  Zeyu Cui and
                  Zhenru Zhang and
                  Zihan Qiu},
  title        = {Qwen2.5 Technical Report},
  journal      = {CoRR},
  volume       = {abs/2412.15115},
  year         = {2024},
  url          = {https://doi.org/10.48550/arXiv.2412.15115},
  doi          = {10.48550/ARXIV.2412.15115},
  eprinttype    = {arXiv},
  eprint       = {2412.15115},
  bibsource    = {dblp computer science bibliography, https://dblp.org}
}

@inproceedings{Tang2025UnlockingGL,
  author       = {Xinyu Tang and
                  Xiaolei Wang and
                  Zhihao Lv and
                  Yingqian Min and
                  Xin Zhao and
                  Binbin Hu and
                  Ziqi Liu and
                  Zhiqiang Zhang},
  editor       = {Wanxiang Che and
                  Joyce Nabende and
                  Ekaterina Shutova and
                  Mohammad Taher Pilehvar},
  title        = {Unlocking General Long Chain-of-Thought Reasoning Capabilities of
                  Large Language Models via Representation Engineering},
  booktitle    = {Proceedings of the 63rd Annual Meeting of the Association for Computational
                  Linguistics (Volume 1: Long Papers), {ACL} 2025, Vienna, Austria,
                  July 27 - August 1, 2025},
  pages        = {6832--6849},
  publisher    = {Association for Computational Linguistics},
  year         = {2025},
  url          = {https://aclanthology.org/2025.acl-long.339/},
  bibsource    = {dblp computer science bibliography, https://dblp.org}
}

@misc{qwq32b,
    title = {QwQ-32B: Embracing the Power of Reinforcement Learning},
    url = {https://qwenlm.github.io/blog/qwq-32b/},
    author = {Qwen Team},
    month = {March},
    year = {2025}
}

@article{Bai2023QwenTR,
  author       = {Jinze Bai and
                  Shuai Bai and
                  Yunfei Chu and
                  Zeyu Cui and
                  Kai Dang and
                  Xiaodong Deng and
                  Yang Fan and
                  Wenbin Ge and
                  Yu Han and
                  Fei Huang and
                  Binyuan Hui and
                  Luo Ji and
                  Mei Li and
                  Junyang Lin and
                  Runji Lin and
                  Dayiheng Liu and
                  Gao Liu and
                  Chengqiang Lu and
                  Keming Lu and
                  Jianxin Ma and
                  Rui Men and
                  Xingzhang Ren and
                  Xuancheng Ren and
                  Chuanqi Tan and
                  Sinan Tan and
                  Jianhong Tu and
                  Peng Wang and
                  Shijie Wang and
                  Wei Wang and
                  Shengguang Wu and
                  Benfeng Xu and
                  Jin Xu and
                  An Yang and
                  Hao Yang and
                  Jian Yang and
                  Shusheng Yang and
                  Yang Yao and
                  Bowen Yu and
                  Hongyi Yuan and
                  Zheng Yuan and
                  Jianwei Zhang and
                  Xingxuan Zhang and
                  Yichang Zhang and
                  Zhenru Zhang and
                  Chang Zhou and
                  Jingren Zhou and
                  Xiaohuan Zhou and
                  Tianhang Zhu},
  title        = {Qwen Technical Report},
  journal      = {CoRR},
  volume       = {abs/2309.16609},
  year         = {2023},
  url          = {https://doi.org/10.48550/arXiv.2309.16609},
  doi          = {10.48550/ARXIV.2309.16609},
  eprinttype    = {arXiv},
  eprint       = {2309.16609},
  bibsource    = {dblp computer science bibliography, https://dblp.org}
}

@inproceedings{Wei2022ChainOT,
  author       = {Jason Wei and
                  Xuezhi Wang and
                  Dale Schuurmans and
                  Maarten Bosma and
                  Brian Ichter and
                  Fei Xia and
                  Ed H. Chi and
                  Quoc V. Le and
                  Denny Zhou},
  editor       = {Sanmi Koyejo and
                  S. Mohamed and
                  A. Agarwal and
                  Danielle Belgrave and
                  K. Cho and
                  A. Oh},
  title        = {Chain-of-Thought Prompting Elicits Reasoning in Large Language Models},
  booktitle    = {Advances in Neural Information Processing Systems 35: Annual Conference
                  on Neural Information Processing Systems 2022, NeurIPS 2022, New Orleans,
                  LA, USA, November 28 - December 9, 2022},
  year         = {2022},
  url          = {http://papers.nips.cc/paper\_files/paper/2022/hash/9d5609613524ecf4f15af0f7b31abca4-Abstract-Conference.html},
  bibsource    = {dblp computer science bibliography, https://dblp.org}
}

@inproceedings{Ye2024PhysicsOL,
  author       = {Tian Ye and
                  Zicheng Xu and
                  Yuanzhi Li and
                  Zeyuan Allen{-}Zhu},
  title        = {Physics of Language Models: Part 2.2, How to Learn From Mistakes on
                  Grade-School Math Problems},
  booktitle    = {The Thirteenth International Conference on Learning Representations,
                  {ICLR} 2025, Singapore, April 24-28, 2025},
  publisher    = {OpenReview.net},
  year         = {2025},
  url          = {https://openreview.net/forum?id=zpDGwcmMV4},
  bibsource    = {dblp computer science bibliography, https://dblp.org}
}

@inproceedings{martschat-markert-2017-improving,
  author       = {Sebastian Martschat and
                  Katja Markert},
  editor       = {Mirella Lapata and
                  Phil Blunsom and
                  Alexander Koller},
  title        = {Improving {ROUGE} for Timeline Summarization},
  booktitle    = {Proceedings of the 15th Conference of the European Chapter of the
                  Association for Computational Linguistics, {EACL} 2017, Valencia,
                  Spain, April 3-7, 2017, Volume 2: Short Papers},
  pages        = {285--290},
  publisher    = {Association for Computational Linguistics},
  year         = {2017},
  url          = {https://doi.org/10.18653/v1/e17-2046},
  doi          = {10.18653/V1/E17-2046},
  bibsource    = {dblp computer science bibliography, https://dblp.org}
}

@inproceedings{Kojima2022LargeLM,
  author       = {Takeshi Kojima and
                  Shixiang Shane Gu and
                  Machel Reid and
                  Yutaka Matsuo and
                  Yusuke Iwasawa},
  editor       = {Sanmi Koyejo and
                  S. Mohamed and
                  A. Agarwal and
                  Danielle Belgrave and
                  K. Cho and
                  A. Oh},
  title        = {Large Language Models are Zero-Shot Reasoners},
  booktitle    = {Advances in Neural Information Processing Systems 35: Annual Conference
                  on Neural Information Processing Systems 2022, NeurIPS 2022, New Orleans,
                  LA, USA, November 28 - December 9, 2022},
  year         = {2022},
  url          = {http://papers.nips.cc/paper\_files/paper/2022/hash/8bb0d291acd4acf06ef112099c16f326-Abstract-Conference.html},
  bibsource    = {dblp computer science bibliography, https://dblp.org}
}

@inproceedings{martschat-markert-2018-temporally,
  author       = {Sebastian Martschat and
                  Katja Markert},
  editor       = {Anna Korhonen and
                  Ivan Titov},
  title        = {A Temporally Sensitive Submodularity Framework for Timeline Summarization},
  booktitle    = {Proceedings of the 22nd Conference on Computational Natural Language
                  Learning, CoNLL 2018, Brussels, Belgium, October 31 - November 1,
                  2018},
  pages        = {230--240},
  publisher    = {Association for Computational Linguistics},
  year         = {2018},
  url          = {https://doi.org/10.18653/v1/k18-1023},
  doi          = {10.18653/V1/K18-1023},
  bibsource    = {dblp computer science bibliography, https://dblp.org}
}

@article{webthinker,
  author       = {Xiaoxi Li and
                  Jiajie Jin and
                  Guanting Dong and
                  Hongjin Qian and
                  Yutao Zhu and
                  Yongkang Wu and
                  Ji{-}Rong Wen and
                  Zhicheng Dou},
  title        = {WebThinker: Empowering Large Reasoning Models with Deep Research Capability},
  journal      = {CoRR},
  volume       = {abs/2504.21776},
  year         = {2025},
  url          = {https://doi.org/10.48550/arXiv.2504.21776},
  doi          = {10.48550/ARXIV.2504.21776},
  eprinttype    = {arXiv},
  eprint       = {2504.21776},
  bibsource    = {dblp computer science bibliography, https://dblp.org}
}

@article{webwaver,
  author       = {Zijian Li and Xin Guan and Bo Zhang and Shen Huang and Houquan Zhou and Shaopeng Lai and Ming Yan and Yong Jiang and Pengjun Xie and Fei Huang and Jun Zhang and Jingren Zhou},
  title        = {WebWeaver: Structuring Web-Scale Evidence with Dynamic Outlines for Open-Ended Deep Research},
  journal      = {CoRR},
  volume       = {abs/2509.13312},
  year         = {2025},
  url          = {https://doi.org/10.48550/arXiv.2509.13312},
  doi          = {10.48550/ARXIV.2509.13312},
  eprinttype   = {arXiv},
  eprint       = {2509.13312},
  bibsource    = {dblp computer science bibliography, https://dblp.org}
}

@article{tongyideepresearch,
  author       = {Baixuan Li and Bo Zhang and Dingchu Zhang and Fei Huang and Guangyu Li and Guoxin Chen and Huifeng Yin and Jialong Wu and Jingren Zhou and Kuan Li and Liangcai Su and Litu Ou and Liwen Zhang and Pengjun Xie and Rui Ye and Wenbiao Yin and Xinmiao Yu and Xinyu Wang and Xixi Wu and Xuanzhong Chen and Yida Zhao and Zhen Zhang and Zhengwei Tao and Zhongwang Zhang and Zile Qiao and Chenxi Wang and Donglei Yu and Gang Fu and Haiyang Shen and Jiayin Yang and Jun Lin and Junkai Zhang and Kui Zeng and Li Yang and Hailong Yin and Maojia Song and Ming Yan and Peng Xia and Qian Xiao and Rui Min and Ruixue Ding and Runnan Fang and Shaowei Chen and Shen Huang and Shihang Wang and Shihao Cai and Weizhou Shen and Xiaobin Wang and Xin Guan and Xinyu Geng and Yingcheng Shi and Yuning Wu and Zhuo Chen and Zijian Li and Yong Jiang},
  title        = {Tongyi DeepResearch Technical Report},
  journal      = {CoRR},
  volume       = {abs/2510.24701},
  year         = {2025},
  url          = {https://doi.org/10.48550/arXiv.2510.24701},
  doi          = {10.48550/ARXIV.2510.24701},
  eprinttype   = {arXiv},
  eprint       = {2510.24701},
  bibsource    = {dblp computer science bibliography, https://dblp.org}
}


\end{document}